\documentclass{article}

\usepackage{arxiv}

\usepackage[utf8]{inputenc} 
\usepackage[T1]{fontenc}    
\usepackage{hyperref}       
\usepackage{url}            
\usepackage{booktabs}       
\usepackage{amsfonts}       
\usepackage{nicefrac}       
\usepackage{microtype}      
\usepackage{lipsum}
\usepackage{graphicx}
\usepackage{wrapfig}
\usepackage{amsmath}
\graphicspath{ {./images/} }

\title{Leveraging Multi-AI Agents for Cross-Domain Knowledge Discovery}

\author{
 Shiva Aryal \\
  Department of Biomedical Engineering\\
  University of South Dakota\\
  Vermillion, SD 57069 \\
  \texttt{shiva.aryal@coyotes.usd.edu} \\
   \And
 Tuyen Do \\
  Department of Biomedical Engineering\\
  University of South Dakota\\
  Vermillion, SD 57069 \\
  \texttt{tuyen.do@usd.edu} \\
  \And
 Bisesh Heyojoo \\
  Department of Biomedical Engineering\\
  University of South Dakota\\
  Vermillion, SD 57069 \\
  \texttt{bisesh.heyojoo@coyotes.usd.edu} \\
  \And
 Sandeep Chataut \\
  Department of Biomedical Engineering\\
  University of South Dakota\\
  Vermillion, SD 57069 \\
  \texttt{sandeep.chataut@coyotes.usd.edu} \\
  \And
 Bichar Dip Shrestha Gurung \\
  Department of Biomedical Engineering\\
  University of South Dakota\\
  Vermillion, SD 57069 \\
\texttt{bichar.shresthagurun@coyotes.usd.edu} \\
  \And
 Venkataramana Gadhamshetty \\
  Dept. of Civil and Environmental Engineering\\
  South Dakota School of Mines and Technology\\
  Rapid City, SD 57701 \\
  \texttt{venkataramana.gadhamshetty@sdsmt.edu} \\
  \And
 Etienne Gnimpieba \\
  Department of Biomedical Engineering\\
  University of South Dakota\\
  Vermillion, SD 57069 \\
  \texttt{etienne.gnimpieba@usd.edu} \\
}

\begin{document}
\maketitle
\begin{abstract}
In the rapidly evolving field of artificial intelligence, the ability to harness and integrate knowledge across various domains stands as a paramount challenge and opportunity. This study introduces a novel approach to cross-domain knowledge discovery through the deployment of multi-AI agents, each specialized in distinct knowledge domains. These AI agents, designed to function as domain-specific experts, collaborate in a unified framework to synthesize and provide comprehensive insights that transcend the limitations of single-domain expertise. By facilitating seamless interaction among these agents, our platform aims to leverage the unique strengths and perspectives of each, thereby enhancing the process of knowledge discovery and decision-making. We present a comparative analysis of the different multi-agent workflow scenarios evaluating their performance in terms of efficiency, accuracy, and the breadth of knowledge integration. Through a series of experiments involving complex, interdisciplinary queries, our findings demonstrate the superior capability of domain specific multi-AI agent system in identifying and bridging knowledge gaps. This research not only underscores the significance of collaborative AI in driving innovation but also sets the stage for future advancements in AI-driven, cross-disciplinary research and application. Our methods were evaluated on a small pilot data and it showed a trend we expected, if we increase the amount of data we custom train the agents, the trend is expected to be more smooth. 
\end{abstract}


\section{Introduction}
The realm of artificial intelligence (AI) has undergone remarkable transformations since its inception, evolving from rudimentary computational algorithms to sophisticated systems capable of mimicking human-like cognitive functions. This rapid advancement has propelled AI into the forefront of technological innovation, making it integral in addressing complex challenges across diverse fields.\cite{Do2023}\cite{Bom2023}\cite{chataut2024comparative} However, as the scope of AI applications broadens, the necessity for cross-domain knowledge discovery emerges as a critical endeavor \cite{Raza2024, Cheng2024}. Traditional AI systems, while adept within their specific areas of expertise, often falter when tasked with synthesizing information across disparate domains. This limitation not only hampers the potential for innovation but also restricts the depth of analysis and understanding that can be achieved. The burgeoning field of AI now stands at a juncture where the integration of knowledge from various disciplines is not just advantageous but essential for pushing the boundaries of what AI can accomplish.

The integration of cross-domain knowledge poses significant challenges in the current landscape of artificial intelligence \cite{Rizvi2023}. Traditional AI models are typically designed with a narrow focus, excelling in tasks within their domain but lacking the capacity to interpret and utilize information beyond their programmed expertise. This siloed approach to knowledge processing leads to a compartmentalization of insights, preventing the holistic understanding necessary for tackling complex, multifaceted problems. Moreover, the task of merging diverse knowledge bases involves intricate considerations of context, relevance, and applicability, areas where conventional AI systems often fall short. As a result, there exists a palpable gap in the ability of existing AI models to seamlessly interact and integrate insights across different domains, limiting their effectiveness in scenarios where interdisciplinary analysis is paramount. The challenge, therefore, lies not only in enhancing the individual capabilities of AI agents but also in fostering a collaborative ecosystem where these agents can collectively leverage their domain-specific expertise for enriched knowledge discovery and decision-making.

We have used MetaGPT for most of our workflows in this research paper. MetaGPT is introduced, an innovative framework that incorporates efficient human workflows as a meta programming approach into LLM-based multi-agent collaboration and leverages the assembly line paradigm to assign diverse roles to various agents, thereby establishing a framework that can effectively and cohesively deconstruct complex multi- agent collaborative problems \cite{Hong2023}. Heree, We can specify the orders of generating answers by the agents. The information passed onto the next agent is the complete context of the results of previous agents which is managed by MetaGPT internally.

The primary aim of this research is to explore the capabilities of existing multi-AI agent platforms for enhancing cross-domain knowledge discovery. This exploration seeks to understand how the orchestration of AI agents, each with domain-specific expertise, can collaboratively tackle complex problems that individual AI capabilities cannot address alone. The research focuses on assessing the efficiency, accuracy, and breadth of knowledge integration facilitated by various multi-AI agent configurations. The goal is to highlight the potential of leveraging collective intelligence among AI agents, marking a significant shift towards more integrative and holistic approaches in cross-disciplinary research and applications.

This research is poised to make significant contributions to the field of artificial intelligence, especially in the realm of collaborative AI technologies. By analyzing the integration capabilities of existing multi-AI agent platforms, it addresses a vital gap in the current AI paradigm: the effective synthesis of domain-specific knowledge from multiple AI agents. The study's contributions are multifaceted. It offers an in-depth evaluation of the potential and limitations of multi-agent AI systems in cross-domain knowledge discovery. It also enriches the theoretical foundations of AI integration, providing insights into the mechanics of effective AI collaboration. Moreover, by showcasing the practical applications and potential benefits of multi-agent AI systems, the research lays the foundation for future advancements in AI-driven, interdisciplinary research and problem-solving. Through these efforts, the study not only advances the understanding of collaborative AI but also illustrates the importance of leveraging AI technologies for comprehensive and inclusive approaches to addressing complex global challenges.

\section{Methodology}
\subsection{Agent Architecture}
The architecture of our multi-AI agent platform is designed to facilitate seamless integration and collaboration between specialized ReAct AI agents \cite{Yao2022}. As depicted in the first figure, each agent within the architecture has a specific role:
\begin{itemize}
    \item observe: The agents begin by observing the environment or the data, gathering information that is relevant to their domain expertise.
    \item think: Each agent processes the observed information, using its domain-specific knowledge to analyze and interpret the data.
    \item act: Based on the analysis, the agent decides on a course of action and executes it, contributing to the problem-solving process.
\end{itemize}

\begin{wrapfigure}{r}{0.5\textwidth}
    \centering
    \includegraphics[width=0.5\linewidth]{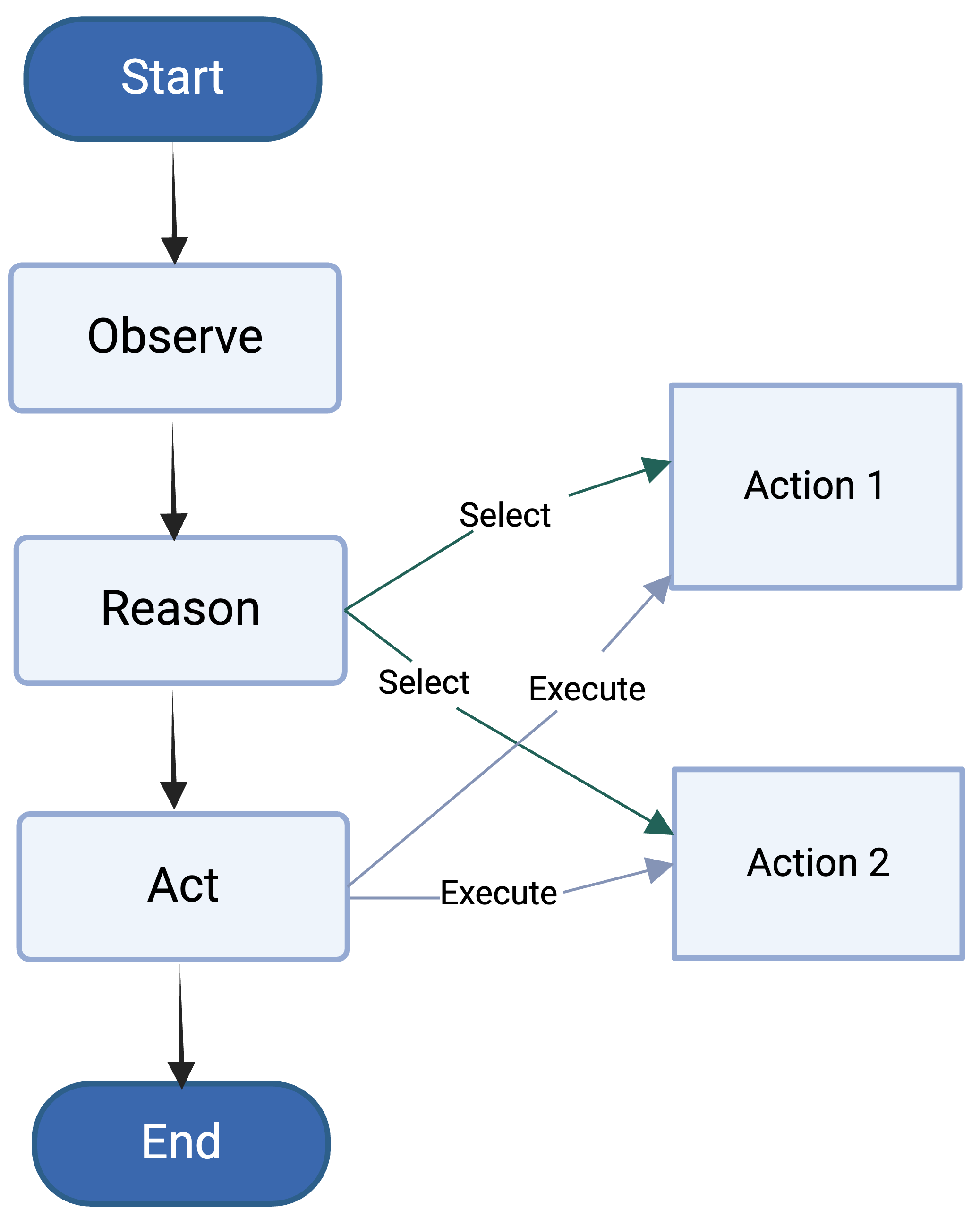}
    \caption{Single Agent Architecure}
\end{wrapfigure}

In creating an AI agent, we begin by defining its knowledge domain and designing algorithms tailored to that domain. Each agent is trained on a dataset consisting of approximately 1000 research papers relevant to its expertise, integrating this specialized knowledge with the broader OpenAI knowledge base.

\subsubsection{Types of Agents Incorporated}
The second figure showcases the five distinct types of AI agents integrated into the multi-agent system:
\begin{itemize}
    \item Boron Nitride Agent
    \item Electrochemical Agent
    \item Bandgap (Physics) Agent
    \item Nanomaterial Agent
    \item AI Agent
\end{itemize}
Each agent has been developed with deep domain knowledge in its respective field, acquired from extensive research literature.

\subsection{Collaboration Mechanism}
Our multi-agent system employs a sophisticated collaboration mechanism, where AI agents communicate and share knowledge via a shared platform. The communication protocol ensures that the knowledge transfer between agents is contextually relevant and precise. The agents collaborate on problem-solving tasks by passing insights and processed information through a well-defined sequence that optimizes the collective intelligence of the system. We have used MetaGPT multi-agent orchestrator that internally passes the context of the entire previous results of agents previously in the MAS system on to the next agent in the sequence.

\subsection{Implementation Details}
The multi-agent system utilizes four distinct flows for knowledge integration:

\begin{figure}
    \centering
    \includegraphics[width=1\linewidth]{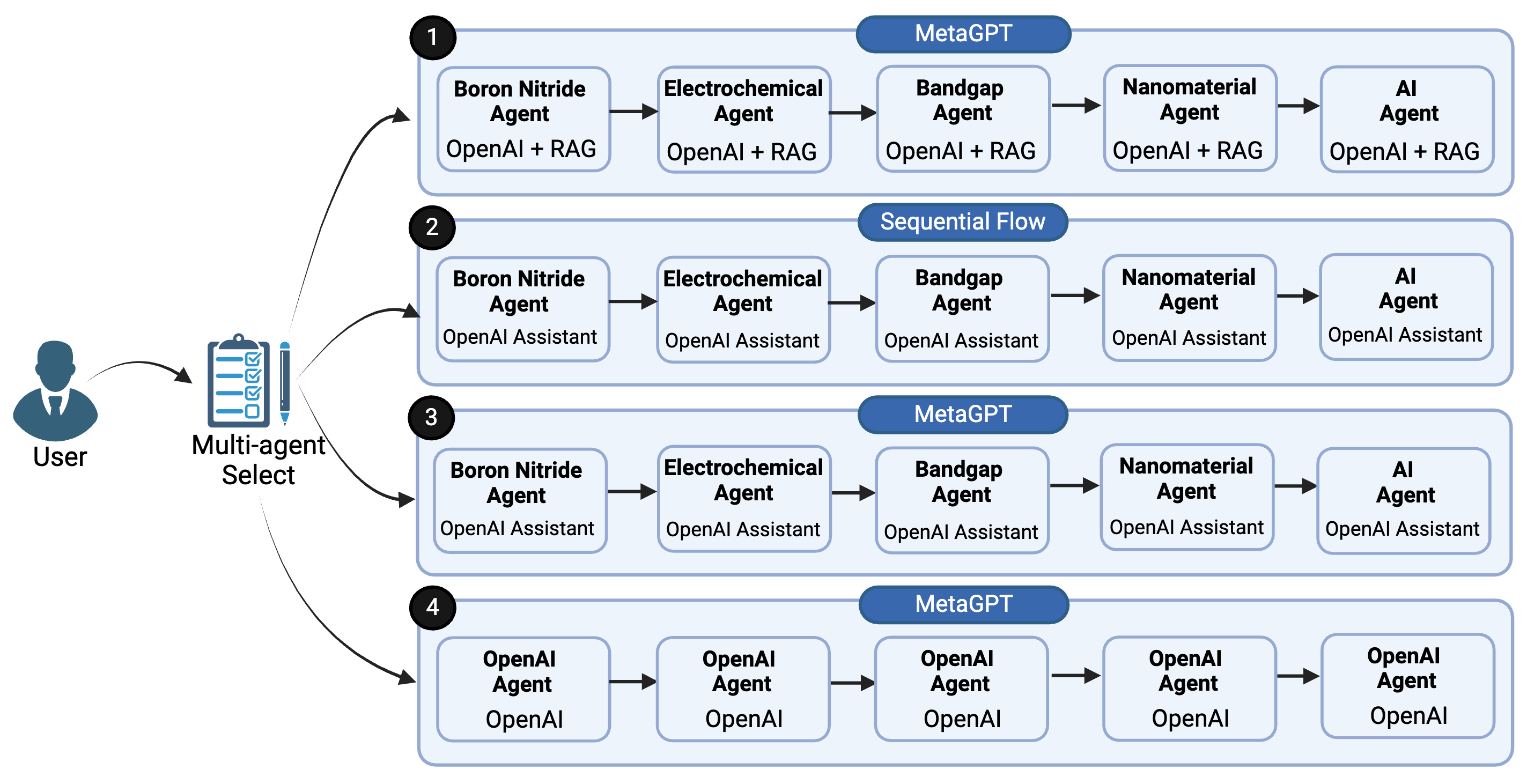}
    \caption{Multi-agent Flows}
\end{figure}

\begin{itemize}
    \item \textbf{MetaGPT+OpenAI+RAG}: A RAG System to create a single agent that augments OpenAI's knowledge with custom data from research papers, orchestrated by the MetaGPT framework.
    \begin{figure}
        \centering
        \includegraphics[width=0.7\linewidth]{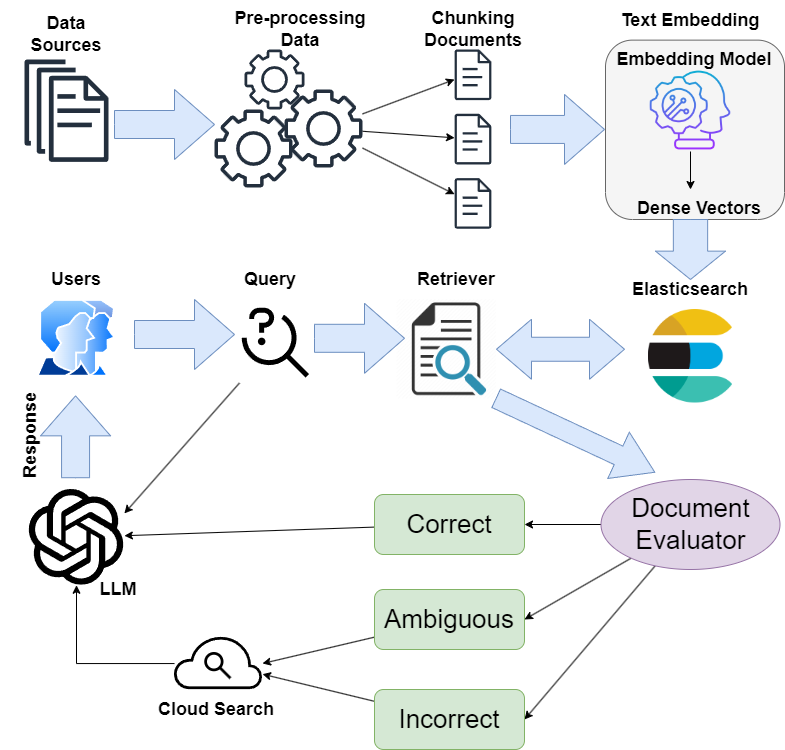}
        \caption{RAG Architecture}
    \end{figure}
    Here individual agent is generated by incorporating custom knowledge from domain specific sources into a generative model. The process begins with a user query, which is sent to a retriever module. This retriever is likely powered by Elasticsearch, and its role is to find relevant documents from a customized knowledge source. These articles are pre-processed through a step called "Chunking documents", which might be to divide the text into manageable pieces. The retriever creates dense vectors using an embedding model, which represents the text chunks numerically in a way that captures semantic meaning.

    The retrieved documents are then evaluated: they can be categorized as 'Correct', 'Ambiguous', or 'Incorrect' based on their relevance to the query. For 'Ambiguous' or 'Incorrect' retrievals, a web search is initiated to augment the data. Finally, the 'User Query' along with 'Context' and additional 'Prompt' information are fed into a OpenAI GPT model. This model generates the output based on both the initial user query and the augmented information retrieved from various sources, thereby integrating custom knowledge into the generative process. The response is then formulated and presented to the user. The process flows is shown in Figure 3.
    \item \textbf{Sequential Flow + OpenAI Assistant}:  A sequential integration of OpenAI Assistant API - based agents, each enriched with domain-specific knowledge from research papers, without the use of an additional orchestration framework, passing the message of immediate past agent to the next agent in the sequence. Here, indiviudal agent is created from OpenAI assistant whose capabilities are expanded through retrieval, incorporating external knowledge beyond its inherent model. This can include proprietary product data or documents supplied by users.
    
    When a document is uploaded and transmitted to the Assistant, OpenAI will automatically chunk the document it into smaller segments, create an index, store the embeddings, and utilise vector search to fetch pertinent information for responding to user inquiries.
    
    \item \textbf{MetaGPT + OpenAI Assistant}: An integration similar to the second flow, but using the MetaGPT orchestration framework to facilitate the interaction between the created agents.
    
    \item \textbf{MetaGPT + OpenAI}: A baseline flow that employs the unmodified OpenAI agent in sequence, serving as a control for assessing the performance enhancement brought by custom knowledge integration.
    
\end{itemize}

\subsection{Comparison Criteria}
To evaluate the efficacy of our multi-agent system against existing platforms, we established a set of metrics and benchmarks. These include:

\begin{itemize}
    \item Efficiency: Measuring the time taken to reach a solution or provide an insight.
    \item Accuracy: Evaluating the correctness and relevance of the information provided by the agents.
    \item Breadth of Knowledge Integration: Assessing the diversity of information and the depth of the synthesized knowledge from multiple domains.
\end{itemize}

These metrics provide a quantitative basis for comparing our proposed system with existing AI agents and platforms, highlighting the improvements in cross-domain knowledge discovery and application.

\section{Experiments and Results}
\subsection{Experimental Setup}
The experimental setup was meticulously designed to test the effectiveness of the multi-AI agent system in delivering accurate and relevant information across different domains. The setup involved the following components:

\begin{itemize}
    \item Questions: We formulated a set of questions within the specific expertise of Boron Nitride, Electrochemical, Bandgap, Nanomaterial, and General AI—each question crafted to assess the depth and breadth of knowledge that the respective agent could access and interpret.
    \item Expected Answers: For each question, we established a set of expected answers that would be considered accurate and comprehensive, serving as a benchmark for the agents' performance.
    \item Test Design Rationale: The reason for this particular design was to emulate realistic scenarios where domain-specific expert knowledge is crucial. The questions were chosen for their impact and relevance in their respective fields to evaluate whether the AI agents could replicate the level of insight expected from human experts.
\end{itemize}

\begin{figure}
    \centering
    \includegraphics[width=0.9\linewidth]{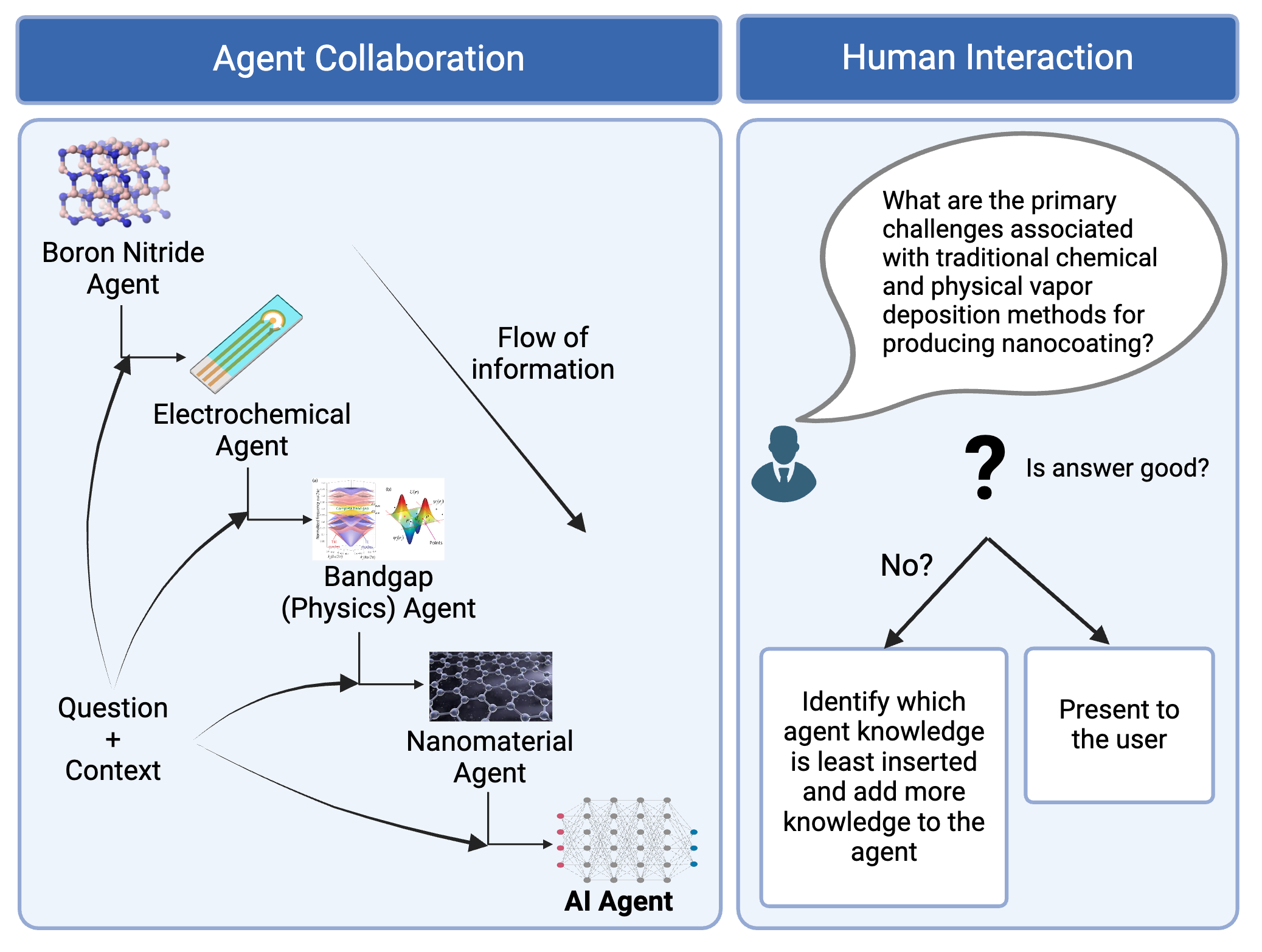}
    \caption{Experimental Setup}
\end{figure}

\subsection{Evaluating Performance}
The performance of the multi-agent system was assessed through two main criteria:

\begin{itemize}
    \item \textbf{Working Performance}: We measured the speed of the system in tokens per second, calculating the time taken from receiving the question to generating the final answer. This metric gauged the efficiency of the system under different workflows.
    \item \textbf{ROUGE}: ROUGE stands for Recall-Oriented Understudy for Gisting Evaluation. This is a really popular metric that you'll definitely find in the literature around text summarization. The metric is based on calculating the syntactic overlap between candidate and reference summaries (or any other text pieces). Rouge-1 calculates the overlap of unigram(individual word) between the candidate and reference text pieces. The mathematical formulation of Rouge-1 precision, recall is shown below -
    \begin{align*}
    R1\text{-recall} & \longrightarrow \frac{\text{number\_of\_overlapping\_words}}{\text{total\_words\_in\_reference\_summary}} \\
    R1\text{-precision} & \longrightarrow \frac{\text{number\_of\_overlapping\_words}}{\text{total\_words\_in\_system\_summary}}
    \end{align*}

    \item \textbf{Cosine Similarity}: 
    Cosine similarity is a metric used to measure how similar two vectors are irrespective of their size. Mathematically, it calculates the cosine of the angle between two vectors projected in a multi-dimensional space. This similarity is particularly useful in high-dimensional positive spaces like those used in text analysis and information retrieval.
    \[
    \text{Cosine Similarity}(\mathbf{A}, \mathbf{B}) = \frac{\mathbf{A} \cdot \mathbf{B}}{\|\mathbf{A}\| \|\mathbf{B}\|}
    \]
    
    \text{Where:}
    \begin{itemize}
        \item $\mathbf{A}$ and $\mathbf{B}$ are the vector representations of two documents (or sentences in this case).
        \item $\mathbf{A} \cdot \mathbf{B}$ is the dot product of vectors $\mathbf{A}$ and $\mathbf{B}$.
        \item $\|\mathbf{A}\|$ and $\|\mathbf{B}\|$ are the norms (or magnitudes) of the vectors, calculated as the square root of the sum of the squared components of each vector.
    \end{itemize}

\end{itemize}

\subsection{Comparative Analysis}
The comparative analysis involved contrasting the performance metrics across the different multi-agent workflows. This included:





\textbf{Speed of Answer Generation}: Comparing the tokens per second across workflows to determine which configuration provided the fastest responses, the average speed of flow 1 was 8.53 tokens per second, flow 2 was 7.63 tokens per second, flow 3 was 8.50 tokens per second whereas flow 4 was 64.23 tokens per second.

\textbf{ROUGE-1}: Comparing the precision across workflows to determine which configuration provided the fastest responses, the average precision of flow 1 was 0.49, flow 2 was 0.05, flow 3 was 0.05 whereas flow 4 was 0.06.


\textbf{Cosine Similarity}:
Comparing the cosine similarity, the average cosine similarity value of flow 1 was 0.26 , flow 2 was 0.22, flow 3 was 0.22 whereas flow 4 was 0.25.

\begin{figure}
    \centering
    \includegraphics[width=1\linewidth]{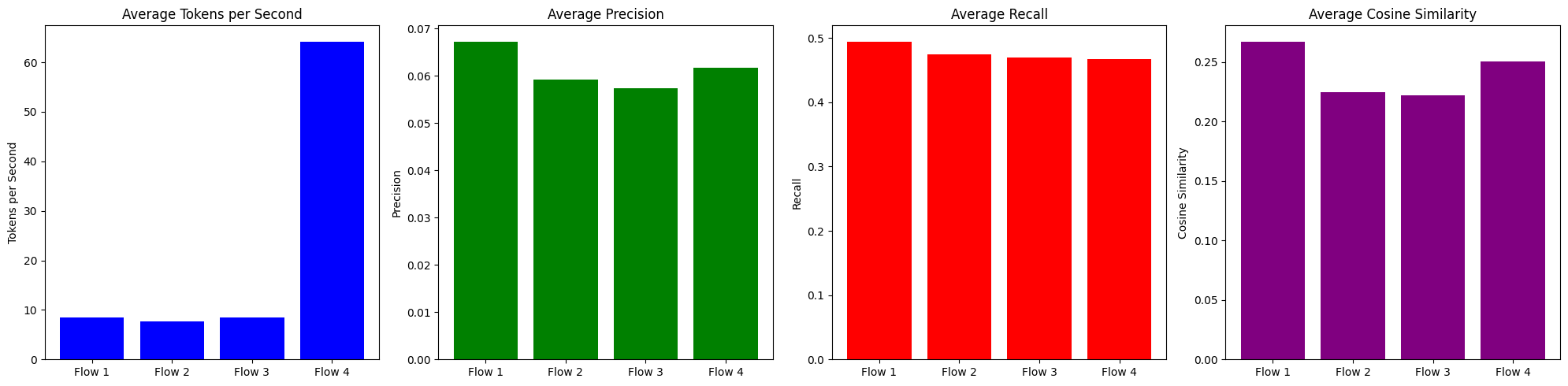}
    \caption{Average Evaluation Metrics}
\end{figure}

We discovered that the different multi-agent workflows exhibited varying strengths. Some workflows were faster but less accurate, while others took longer but provided more comprehensive answers. The expert evaluations were instrumental in understanding the practical utility of each system configuration. They highlighted that the most impactful questions often required not just speed but depth of analysis, a quality that some workflows managed better than others. The results from these experiments provide valuable insights into how to balance efficiency and thoroughness in the design of multi-agent AI systems.

\section{Discussion}
Our study marks a transformative advancement in artificial intelligence, with a particular emphasis on the fusion of cross-domain knowledge discovery and integration. Through the deployment of a sophisticated network of multi-AI agents, each expert in distinct knowledge areas, we uncover the vast potential of collaborative AI systems equipped with the capability to harness custom tools and open data. This innovative strategy not only dismantles the traditional barriers that have impeded AI applications but also heralds a new era of intelligent systems designed for evolution and adaptation to serve a wide array of research disciplines.

The remarkable ability of our AI agents to efficiently and accurately synthesize knowledge from diverse domains, as demonstrated by our experiments, showcases the significant advantages of utilizing domain-specific expertise through a unified, collaborative framework. Such enhanced performance is largely due to the dynamic interplay and cooperation among the AI agents. This allows for a deeper and more nuanced comprehension of complex queries, enriched further by the agents' ability to access and utilize custom tools and open data sources, thus broadening the scope and depth of their analytical capabilities.

A pivotal aspect of our findings is the AI agents' unique capability not just to collaborate but also to innovate and develop new AI models specifically tailored to address intricate research challenges. This introduces an unprecedented level of adaptability and evolutionary potential within AI systems. As these agents craft and refine novel models, their problem-solving methods are continually enhanced, leading to a rapid acceleration in knowledge discovery and practical application.

Despite these encouraging outcomes, we recognize several hurdles, such as the necessity for advanced coordination mechanisms to manage the interactions among AI agents and to ensure the system's scalability as the diversity and number of domains and agents grow. Future endeavors should concentrate on fine-tuning these collaborative dynamics and devising more efficient strategies for the agile creation and implementation of new AI models within the ecosystem, leveraging the full spectrum of custom tools and open data available to them.

\section{Conclusion}

The research conducted provided substantial insights into the effectiveness of multi-AI agent systems in enhancing cross-domain knowledge discovery. The key outcomes indicated that the integration of domain-specific knowledge significantly improves the quality of the AI's output. Flow 1 emerged as the most effective, with experts rating it highest for answer quality. This flow capitalized on the MetaGPT framework's ability to maintain conversational context across all agents, proving essential for delivering high-quality and contextually relevant responses. Flow 3 followed closely, benefiting from the domain-specific knowledge incorporated into the OpenAI Assistant, despite lacking the full conversational context provided in Flow 3. Flow 2, which employed a custom RAG system atop the OpenAI model, demonstrated that while the RAG approach adds value, the OpenAI Assistant's capabilities surpassed it when it comes to answer quality. Flow 4, relying solely on the generalized OpenAI GPT model without additional domain knowledge, lagged in performance, underscoring the importance of domain-specific information in delivering quality AI responses.

The findings from this study open several avenues for future research. Future work could explore the refinement of the MetaGPT orchestration framework to enhance the efficiency and quality of multi-agent collaborations further. Additionally, expanding the domain-specific knowledge databases and integrating real-time learning capabilities could make the system more robust and applicable to a wider range of cross-disciplinary queries. There is also potential in exploring the use of different machine learning models and architectures that could complement the capabilities of GPT-powered agents.

This study has underscored the significant role that collaborative AI can play in advancing knowledge discovery across various disciplines. The integration of domain-specific expertise within AI systems is crucial for tackling complex, multi-faceted problems that are beyond the scope of any single domain. By harnessing the collective intelligence of specialized AI agents, we can open up new possibilities for innovation and understanding, breaking down the barriers that have traditionally segmented knowledge. The importance of collaborative AI will only grow as we continue to push the boundaries of what AI can achieve, and this research lays a critical foundation for those future endeavors.

\end{document}